\documentclass[accepted]{uai2023} 

\usepackage[american]{babel}

\usepackage{natbib} 
\usepackage{mathtools} 
\usepackage{amsfonts}
\usepackage{booktabs} 
\usepackage{tikz} 
\usepackage{algorithm}
\usepackage{algorithmic}

\title{Retrieval-Augmented Meta Learning for Low-Resource Text Classification}

\author[1]{{Rongsheng Li}{}}
\author[1]{{Yangning Li}{}}
\author[1]{{Yinghui Li}{}}
\author[1]{{Chaiyut Luoyichi}{}}
\author[1]{{Hai-Tao Zheng}{}}
\author[2]{{Nannan Zhou}}
\author[2]{{Hanjing Su}}

\affil[1]{%
    Tsinghua University
}
\affil[2]{%
    Wechat Pay\\
    Tencent
}

\begin{document}

\maketitle

\begin{abstract}
Meta learning have achieved promising performance in low-resource text classification which aims to identify target classes with knowledge transferred from source classes with sets of small tasks named episodes. However, due to the limited training data in the meta-learning scenario and the inherent properties of parameterized neural networks, poor generalization performance has become a pressing problem that needs to be addressed. To deal with this issue, we propose a meta-learning based method called Retrieval-Augmented Meta Learning(RAML). It not only uses parameterization for inference but also retrieves non-parametric knowledge from an external corpus to make inferences, which greatly alleviates the problem of poor generalization performance caused by the lack of diverse training data in meta-learning. This method differs from previous models that solely rely on parameters, as it explicitly emphasizes the importance of non-parametric knowledge, aiming to strike a balance between parameterized neural networks and non-parametric knowledge. The model is required to determine which knowledge to access and utilize during inference. Additionally, our multi-view passages fusion network module can effectively and efficiently integrate the retrieved information into low-resource classification task. The extensive experiments demonstrate that RAML significantly outperforms current SOTA low-resource text classification models.

\end{abstract}

\section{Introduction}\label{sec:intro}

Over the last few years, meta-learning have achieved impressive results on low-resource text classification(\citealp{vinyals2016matching}; \citealp{finn2017model}; \citealp{sui2021knowledge}; \citealp{yao2021knowledge}; \citealp{DBLP:journals/patterns/LiuLTLZ22}). The key to success in meta-learning is learning good generalizations from only a small amount of data. However, vanilla meta learning can not generalize well when training tasks belong to limited domains, and features in the query sentence can not be effectively extracted by the models which are training on these training tasks. In the field of natural language (~\citealp{DBLP:conf/acl/LiZLLLSWLCZ22}; \citealp{DBLP:conf/emnlp/MaLSZHZLLLCZS22}), in particular, it becomes important to mitigate the differences between training and testing tasks because of new concepts that may only emerge during testing tasks.

\begin{figure}
    \centering
    \includegraphics[width=1\linewidth]{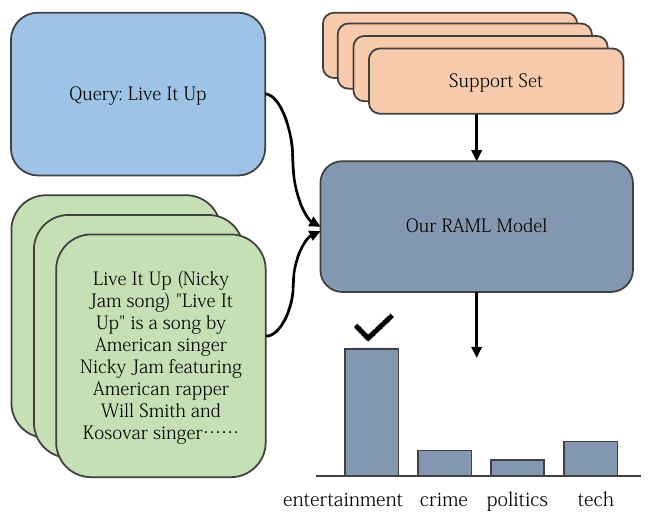}
    \caption{A simple meta-learning method that leverages external information. First, it retrieves supporting text passages from external knowledge sources such as Wikipedia. Then, a passages fusion network based on cross-attention mechanism scores the prototypes using queries and the retrieved passages.}
    \label{fig:intro}
\end{figure}

Meanwhile, some believe that introducing external knowledge may improve the problem of poor generalization performance in meta-learning, because after introducing external knowledge in the training process, the model may be able to learn concepts that would have only appeared during testing. \citet{sui2021knowledge} and \citet{yao2021knowledge} have made attempts to introduce external knowledge into meta-learning, but obtain little performance improvement. Specifically, \citet{yao2021knowledge} attempted to enhance the embedding of the original sentence by learning the embedding of specific entities in the sentence, in order to improve generalization performance. \citet{sui2021knowledge} uses exact string matching to retrieve the relevant KG embeddings from the support set to train a parameter generation network, which is used to initialize the parameters of a metric learning model. The reason why they did not achieve good performance is likely because they only used parameterized networks for training. However, parameterized networks inherently suffer from poor generalization performance, as they may only remember the shallow features of the data.

Recently, models based on text retrieval have been used to improve question answering[\citealp{ijcai2022p0383}; \citealp{wang2022training}], NER and RE[\citealp{zhang2022domain}; \citealp{wang2022named}], machine translation[\citealp{zhang2018guiding}; \citealp{xu2020boosting}]. These retrieval based methods retrieve query-related information from the external knowledge to enhanced the input representation and introduce external knowledge to models, which can greatly improve the generalization performance of the model. Our proposed RAML introduces external knowledge into meta-learning for the first time, the approach inspired by the fact that humans are able to effortlessly learn new concepts from a few examples, largely because they have a knowledge base stored in their brains. For previous models, the learned world knowledge is stored implicitly in the parameters of the underlying neural network. The learned world knowledge in language models is implicitly stored in the parameters of the neural network, making it challenging to determine what knowledge is stored in the network and where. Meanwhile, the size of the network limits storage capacity, requiring larger networks to capture more knowledge, which can be time-consuming and costly.

In this paper, we propose to leverage external knowledge corpus in order to bridge the gap between the training and test tasks and enable more efficient meta-learning for low-resource text classification. This method differs from previous parameter-based models by explicitly highlighting the importance of world knowledge, as it requires the model to determine which knowledge to access and utilize during inference. Prior to each prediction, our model employs a retriever to obtain documents from a vast corpus, such as Wikipedia, and subsequently focuses on those documents to enhance the accuracy of its prediction.

The main contributions of our work are:
\begin{itemize}
    \item We propose a simple but effective method to inject knowledge-aware information into a meta-learning low-resource text classification tasks with retrieval. To our best knowledge, this approach is being proposed for the first time.
    \item A novel multi-view cross-attention module named passages fusion network is proposed to integrate the retrieved information for diverse tasks. This module has the characteristic of semantic preservation, which 
    greatly alleviates the problem of poor generalization performance. More importantly, it can fuse external information retrieved from multiple perspectives.
    \item The extensive experiments results show that our model can effectively integrate the retrieved knowledge and significantly improve the performance of low-resource text classification.
\end{itemize}

\section{Ralated Work}
\paragraph{Text classification via meta learning}
In most modern meta-learning frameworks, two main strategies are widely adopted: optimization-based meta-learning and metric-based meta-learning. All of the optimization-based meta-learning consist of an inner algorithm and an outer algorithm. Examples of optimization-based algorithms include Model-Agnostic Meta-Learning (MAML)\citep{finn2017model} and Reptile\citep{nichol2018reptile}. By leveraging BERT\citep{kenton2019bert}, LEOPARD\citep{bansal2020self} attains strong performance across various classification tasks. Prototypical Network\citep{snell2017prototypical}, Induction Network\citep{geng2019induction}, and Relation Network\citep{sung2018learning} are designed to establish a metric space between classes and samples. Additionally, \citet{bao2020few} have proposed a meta-learning approach that utilizes distributional signatures for few-shot text classification. More recently, KGML\citep{yao2021knowledge} uses extracted sentence-specific entity representations to enhance the representation of the sentence, another one\citep{sui2021knowledge} use the retrieved knowledge to initialize the parameters of the neural network. However, these algorithms suffer from overfitting caused by the imbalance between the few data and the deep model in the few-shot setting and they only employ parameterized networks to make predictions. On the contrary, our RAML model makes predictions using both parameterized neural networks and non-parameterized knowledge retrieved from a large corpus.

\paragraph{Retrieval-based NLP}
For many NLP tasks~\citep{DBLP:journals/corr/abs-2307-09007, DBLP:conf/emnlp/LiMZLLHLLC022}, retrieval is an effective method to utilize external knowledge. Knowledge retrieval has been utilized in a variety of NLP tasks[\citealp{DBLP:conf/sigir/LiLHYS022}; \citealp{DBLP:journals/corr/abs-2207-08087}; ], including question answering[\citealp{ijcai2022p0383}; \citealp{wang2022training}], machine translation[\citealp{zhang2018guiding}; \citealp{xu2020boosting}], NER[\citealp{zhang2022domain}; \citealp{wang2022named}], and entity linking[\citealp{zhang2021entqa}; \citealp{huang2022damo}; \citealp{DBLP:journals/corr/abs-2306-12245}]. Recently, knowledge retrieval has been introduced to language model pretraining. For example, REALM\citep{guu2020retrieval} trains a latent knowledge retriever and a knowledge-augmented encoder in an end-to-end manner during pretraining and fine-tuning. REALM decomposes the generative process into retrieving and predicting, treating the retrieved knowledge as a latent variable and marginalizing it. Our RAML module is inspired by this generative process and also treats retrieved knowledge from text as a latent variable. However, since REALM's retriever is trainable, it requires asynchronous re-embedding and re-indexing of all documents during training. To handle a larger database size, RETRO\citep{borgeaud2022improving} utilizes a frozen retriever along with a chunked cross-attention mechanism to operate on databases containing trillions of tokens. Similarly, in order to optimize efficiency, we also freeze the retriever module in RAML.

\section{Problem Setting}

Few-shot classification is a type of machine learning that focuses on building models that can quickly adapt to new tasks or data sets. Unlike traditional machine learning, which requires large amounts of training data for each individual task, few-shot algorithms are designed to learn how to learn from a smaller set of examples. This is accomplished by training the model on a diverse set of tasks or data sets, which allows it to learn generalizable patterns that can be applied to new problems. The goal of few-shot classification is to create models that can rapidly adapt to new tasks with minimal additional training, making them ideal for applications where new data is constantly being generated. 

In meta-learning setting, few-shot classification algorithms typically involve a two-stage learning process. In the first stage, the model is trained on a set of tasks or data sets, and the goal is to learn a set of parameters or weights that can be used to make predictions on new tasks. This stage is often referred to as meta-training, and the resulting learned parameters are referred to as meta-parameters. In the second stage, the model is fine-tuned on a specific task or data set using the learned meta-parameters. This stage is often referred to as meta-testing, and the goal is to quickly adapt the model to the new task with minimal additional training. The success of few-shot classification depends on the quality and diversity of the meta-training tasks, as well as the ability of the model to learn generalizable patterns that can be applied to new tasks.

More concretely, if the support set contains $N$ labeled examples for each of $C$ unique classes, the few-shot problem is called $C$-way $N$-shot. To ensure good generalization during testing, the $C$-way $N$-shot problem involves episodically sampling support and query sets. In each meta-training iteration, an episode is formed by randomly selecting $C$ classes from the training set with $N$ labeled examples for each of the $C$ classes act as the support set $S={(x_i, y_i)}_{i=1}^{C \times N}$, and a fraction of those $C$ classes' examples serve as the query set $Q={(x_i, y_i)}_{i=1}^{q}$, where $x_i$ is the sentence and $y_i$ is the corresponding label and $Q\cap S = \emptyset$. The model is trained on the support set to minimize loss of predictions over the query set, and this process is repeated iteratively until convergence.

\begin{figure*}
    \centering
    \includegraphics[width=1\linewidth]{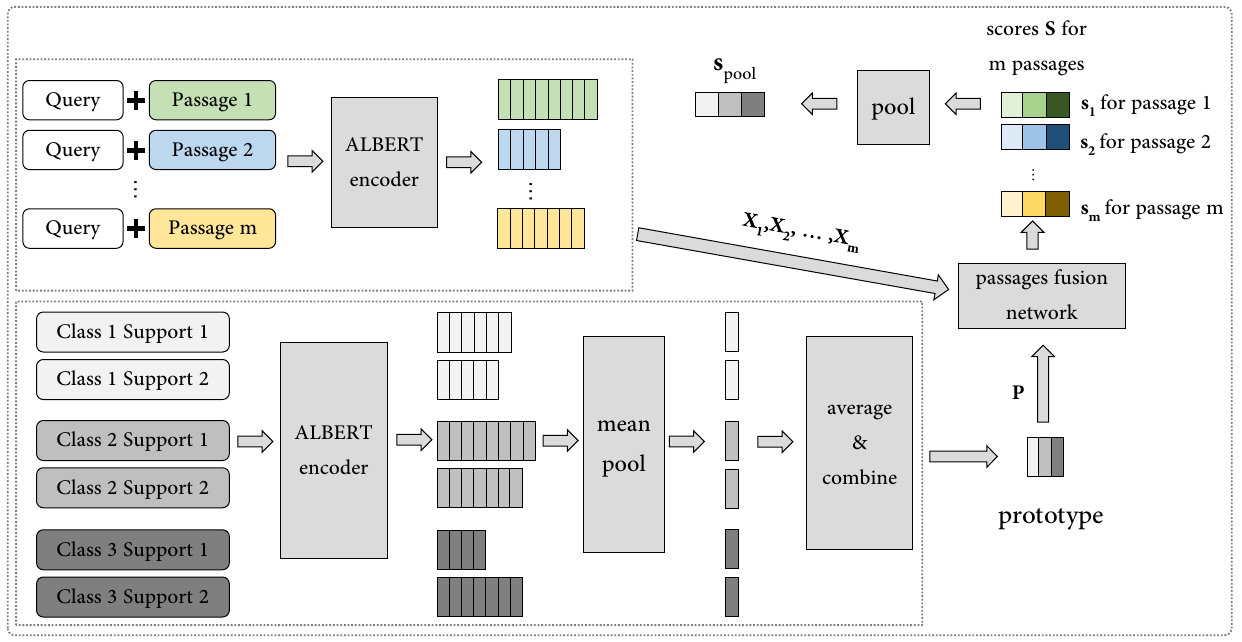}
    \caption{RAML framework for a C-way N-shot (C=3, N=2) problem with one query example. First, we sample task $\mathcal{T}_i \sim p(\mathcal{T})$. For the support set, we have three classes, with two samples in each class. They can be used to generate three prototypes representing their respective classes. For the query set, only one example is shown in the figure, but multiple queries can be computed in parallel. For a single query in the figure, we retrieve $m$ passages, and for the $k$-th passage, we concatenate the tokens corresponding to the query to obtain a matrix $\mathbf{X}_k$. We feed the prototypes $\mathbf{P}$ and $\mathbf{X}_k$ into the passages fusion network to obtain a score, this process is illustrated in Figure~\ref{fig:fusion}.}
    \label{fig:overview}
\end{figure*}

\section{Methodology}
In this section, we present the proposed RAML framework (Figure~\ref{fig:overview}), which allows us to makes predictions using both parameterized neural networks and non-parameterized knowledge retrieved from a large corpus described in the previous section.

\subsection{Sentence Embedding Network}
In this network, a pre-trained ALBERT\citep{lan2019albert} encoder is used to model sentences. Given an input text $x_{i}=([CLS], w_{1}, w_{2}, ..., w_{T} , [SEP])$, the output of ALBERT encoder is denoted as $\mathbf{H}(x_{i}) \in \mathbb{R}^{(T+2) \times d}$, where $d$ is the output dimension of the ALBERT encoder. We denote $j$-th vector of $\mathbf{H}(x_{i})$ by $\mathbf{f}(x_{i}, j)$ and use the mean of all output vectors as the sentence representation, which is denoted as $\mathbf{f}(x_{i}) = \frac{1}{T+2} \sum_{j=1}^{T+2}\mathbf{h}(x_{i}, j) \in \mathbb{R}^{d}$.

In meta-learning, the representation of each class is the mean vector of the embedded sentences belonging to its class,

\begin{equation}\label{1}
\mathbf{p}_{z} = \frac{1}{\left | \mathcal{S}_{z} \right | } \sum_{(x_{i},y_{i}) \in \mathcal{S}_{z}} \mathbf{f}(x_{i}) \in \mathbb{R}^{d}
\end{equation}

where $\mathcal{S}_{z}$ denotes the set of sentences labeled with class $z$. 

\begin{figure}[t]
    \centering
    \includegraphics[width=0.7\linewidth]{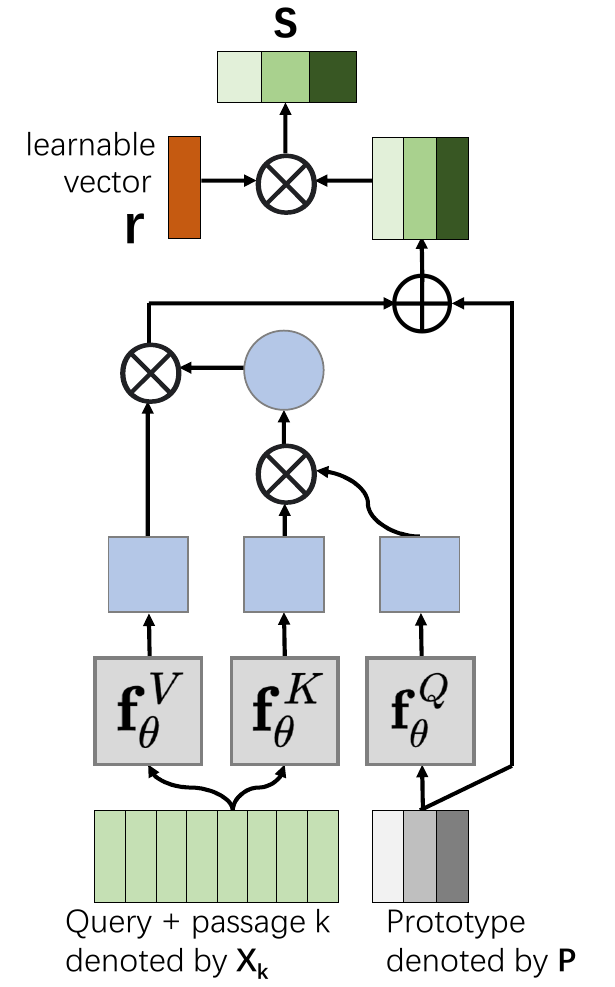}
    \caption{Demonstration of how our passages use a query and a single passage to obtain the final classification score. First, we use the prototype vector $\mathbf{P}$ to obtain the query $\mathbf{Q}$ in cross-attention, and then use the matrix $\mathbf{X}_k$ corresponding to the query and a single passage to derive K and V. After cross-attention, we obtain the vector $\mathbf{Z}$, and by adding the residual $\mathbf{P}$, we obtain $\mathbf{\tilde{Z}}$. Finally, we multiply $\mathbf{\tilde{Z}}$ with the learnable vector $\mathbf{r}$ to obtain the final score for the prototype, corresponding to one view of the prototype. Note that this demonstration is for the case of the number of heads, denoted as h, is equal to 1.}
    \label{fig:fusion}
\end{figure}

\subsection{Multi-View Passages Fusion Network}
As shown in Figure~\ref{fig:overview}, for each query, we retrieve $m$ pieces of auxiliary information, and the retrieval method will be described in detail in the following section. More precisely, each retrieved passage and its title are concatenated with the query, and processed independently from other passages by the encoder. We add special tokens \textit{title:} and \textit{context:} before the title and text of each passage.

As previously discussed in \citet{chen2018zero}, few-shot learning tasks have a tendency to result in semantic loss. This occurs when certain features are discarded during training because they do not provide discrimination for seen classes(belonging to training set), but are crucial for recognizing unseen classes(belonging to testing set). We tackle this problem by applying cross-attention mechanism to assign different weights to each token of query.

The computation process of this subsection is shown in Figure~\ref{fig:fusion}.

 First, queries Q, keys K and values V are computed by applying linear transformations:

\begin{equation}
     \mathbf{Q} = \mathbf{P} \mathbf{W}_{Q}, 	\ \mathbf{K} = \mathbf{X} \mathbf{W}_{K}, 	\ \mathbf{V} = \mathbf{X} \mathbf{W}_{V}.
\end{equation}

where $\mathbf{P} = [\mathbf{p}_{1}, \mathbf{p}_{2}, ... , \mathbf{p}_{c}] \in \mathbb{R}^{c \times d}$ is the prototype of $C$ classes, $\mathbf{X} \in \mathbb{R}^{(T+2) \times d}$ is the embeddings of a query concatenated with one passage. It is worth noting that the $\mathbf{Q}$ notation used here is specifically related to features in the attention module, and should be considered separate from the notion of queries in the few-shot setting.

Then a similarity score between the $i$-th prototype, $\mathbf{Q}_i$, and the $j$-th token of the passage, $\mathbf{K}_j$, is obtained by computing the dot-product between these two elements, and normalized over the dimension:

\begin{equation}
\alpha_{i,j} = \mathbf{Q}_{i}\mathbf{K}_{j}^{T}, \quad \tilde{\alpha}_{i,j}=\frac{\alpha_{i,j}}{ {\sum_{k}\alpha_{i,k}} }
\end{equation}
 
If a vector $K_j$ corresponding to one token that contributes more to the classification, and its label belongs to the class associated with the current prototype $Q_i$, the $\tilde{\alpha}$ will become larger during the non-terminal sample-based iterative training process. This reflects the semantic preservation feature of our module.

Finally, the scores of the classes corresponding to these prototypes $\mathbf{P}$ towards the current query sentense with one passage denoted by $\mathbf{X}$ were calculated:
\begin{equation}    \label{4}
    \begin{split}
\mathbf{Z} &= Cross\_attn(\mathbf{P},\mathbf{X},\mathbf{X})\\
	&= Concat(head_1,head_2, ... ,head_h)\mathbf{W_O}\\
{\rm where} \ head_i &= Softmax(\frac{\mathbf{P} \mathbf{W}_{Q,i} (\mathbf{XW}_{Q,i})^T}{\sqrt{d_h}}) \mathbf{XW}_{V,i}
    \end{split}
\end{equation}
\begin{equation} \label{5}
\mathbf{\tilde{Z}} = \mathbf{Z + P}
\end{equation}
\begin{equation}    \label{6}
\mathbf{s} = \mathbf{\tilde{Z} \times r}
\end{equation}
where $d_h = d/h$, $\mathbf{W}_{Q,i}$,$\mathbf{W}_{K,i}$,$\mathbf{W}_{V,i} \in \mathbb{R}^{d \times d_h}$, $\mathbf{W_O} \in \mathbb{R}^{d \times d}$ are learnable parameters, $\mathbf{r} \in \mathbb{R}^d$ is a learnable vector. The $\mathbf{s} \in \mathbb{R}^c $ is the scores of the current sentence with one passage towards $C$ classes. Meanwhile, for each query, we have $m$ retrieved passages, which means we can obtain $m$ vectors like this simultaneously: $\mathbf{S} = [\mathbf{s}_1, \mathbf{s}_2, ... , \mathbf{s}_m] \in \mathbb{R}^{m \times c}$.

For each passage of the query sentence, we can obtain a vector $\mathbf{s}_i$, which corresponds to a view of the query sentence that we observe. To obtain the final multi-view scores, we adds a pooling operation to the matrix $\mathbf{S}$ to derive the vector $\mathbf{s}_{pool}\in \mathbb{R}^{c}$. We experiment with two pooling strategies: computing the mean of all vectors (\textit{MEAN}-strategy), and computing a max-over-time of the vectors (\textit{MAX}-strategy). The default configuration is \textit{MEAN}. $\mathbf{s}_{pool}$ is calculated as follows:

\begin{equation}    \label{7}
    \mathbf{s}_{pool, MEAN} = \frac{1}{m}\sum_{i=1}^{m}\mathbf{s}_i
\end{equation}

\begin{equation}
    \mathbf{s}_{pool, MAX} = [\max_{i=1}^{m}\mathbf{s}_i^{1}, \max_{i=1}^{m}\mathbf{s}_i^{2},\ ...\ ,\max_{i=1}^{m}\mathbf{s}_i^{c}]
\end{equation}
where $\mathbf{s}_i^{j}$ is the $j$-th value of vector $\mathbf{s}_i$.
\subsection{Knowledge Retrieval}
Following \citet{karpukhin2020dense} and \citet{izacard2021leveraging}, we use the English Wikipedia dump from Dec. 20, 2018 as the source documents for retrieval. We split each article into multiple, disjoint text blocks of 100 words as passages, serving as our basic retrieval units, following (Wang et al., 2019), which results in 21,015,324 passages in the end, each passage is also associated with the title of the Wikipedia article where the passage is from.

Since traditional sparse vector retrieval algorithms, such as TF-IDF or BM25, can not effectively capture the relevant information in a sentence, we use Sentence-BERT\citep{reimers2019sentence} to retrieve passages, which fine-tunes BERT in a siamese/triplet network architecture. In Sentence-BERT, passages and queries are represented as dense vector representations, computed using a single BERT networks. The ranking function is the dot product between the query and passage representations. Then at runtime, passages with the highest similarity score with the input query are retrieved, by using an efficient similarity search library such as FAISS\footnote{https://github.com/facebookresearch/faiss}.

\begin{table*}[htbp]
  \centering
  \caption{Comparison of averaged accuracy on two datasets. The scores of baselines are taken from KGML. We report the averaged accuracy over 600 meta-testing tasks.}
    \begin{tabular}{c|cc|cc}
    \toprule
    Dataset  & \multicolumn{2}{c|}{Amazon Review} & \multicolumn{2}{c}{Huffpost} \\
    Shot  & 1-shot & 5-shot & 1-shot & 5-shot \\
    \midrule
    MAML\citep{finn2017model} & 44.35\% & 56.94\% & 39.95\% & 51.74\% \\
    ProtoNet\citep{snell2017prototypical} & 55.32\% & 73.30\% & 41.72\% & 57.53\% \\
    InductNet\citep{geng2019induction} & 45.35\% & 56.73\% & 41.35\% & 55.96\% \\
    MatchingNet\citealp{vinyals2016matching} & 51.16\% & 69.89\% & 41.18\% & 54.41\% \\
    REGRAB\citep{qu2020few} & 55.07\% & 72.53\% & 42.17\% & 57.66\% \\
    KGML-MAML\citep{yao2021knowledge} & 51.44\% & 58.81\% & 44.29\% & 54.16\% \\
    KGML-ProtoNet\citep{yao2021knowledge} & 58.62\% & 74.55\% & 42.37\% & 58.75\% \\
    \midrule
    RAML(Ours) & \textbf{63.08}\%    & \textbf{75.52}\%    & \textbf{46.62}\% & \textbf{61.53}\% \\
    \bottomrule
    \end{tabular}%
  \label{tab:all}%
\end{table*}%

\subsection{Loss Function}
For each pair $(x_i, y_i)$ in query set $Q={(x_i, y_i)}_{i=1}^{q}$, and its prediction vector $\mathbf{s}_{pool} \in \mathbb{R}^{c}$ upon prototypes corresponding $C$ classes, a Softmax classifier is used:
\begin{equation}    \label{9}
    p(y=y_i|x_i) = \frac{exp(\mathbf{s}_{pool,i})}{\sum_{j=1}^{c}exp(\mathbf{s}_{pool,j})}
\end{equation}
where $\mathbf{s}_{pool,i}$ is the $i$-th value of vector $\mathbf{s}_{pool}$. Finally, the model is trained by minimizing the losses across $N$ episodes:
\begin{equation}    \label{10}
    \mathcal{L} = \frac{1}{N} \sum_{i} \mathcal{L}_i
\end{equation}
where $\mathcal{L}_i$ is the loss of the $i$-th episode over query set $Q$:
\begin{equation}    \label{11}
    \mathcal{L}_i = - \frac{1}{|Q|} \sum_{(x_i,y_i)\in Q} \log{p(y=y_i|x_i)}
\end{equation}
The training process is summarized in Algorithm~\ref{alg1}.

\begin{algorithm}
    
    \caption{RAML training algorithm.}
    \renewcommand{\algorithmicrequire}{\textbf{Input:}}
    \renewcommand{\algorithmicensure}{\textbf{Output:}}
    \label{alg1}
    \begin{algorithmic}[1]
        \REQUIRE{Task distribution $p(\mathcal{T})$; Knowledge Corpus}
        \ENSURE{learned model parameters}
        \STATE Randomly initialize the learnable vector $\mathbf{r}$
        \STATE Initialize ALBERT with albert-base-v1
        \STATE Initialize passages fusion network(without $\mathbf{r}$) with the parameters of ALBERT
        \WHILE {not converge}
        \STATE Sample task $\mathcal{T}_i \sim p(\mathcal{T})$ with support set $S={(x_i, y_i)}_{i=1}^{C \times N}$ and query set $Q={(x_i, y_i)}_{i=1}^{q}$.
        \STATE Get prototype via Eq.(~\ref{1})
        \STATE Get $\mathbf{s}_{pool}$ by Eq.(~\ref{4})\~Eq.(~\ref{7})
        \STATE Update model by Eq.(~\ref{9})\~Eq.(~\ref{11})
        \ENDWHILE
    \end{algorithmic}
\end{algorithm}


\section{Experiment}
In this section, we demonstrate the efficacy of our proposed RAML on two datasets, and carry out a relevant analytical study.

\subsection{Dataset}
We assess the performance of our model on two commonly utilized datasets for text classification. The first dataset is the Amazon Review dataset\citep{ni2019justifying}, which involves assigning categories to reviews. The second dataset is the Huffpost dataset\citep{misra2019sarcasm}, which involves categorizing news headlines. For both the Amazon Review and Huffpost datasets, we employ the conventional C-way N-shot few-shot learning setting\citep{finn2017model}, with C set to 5.

\paragraph{External knowledge corpus}We use the Wikipedia dump from Dec. 20, 2018 for support documents, splitting articles into non-overlapping passages of 100 tokens, and applying the same preprocessing as \citet{chen2017reading}. For computational efficiency, the default number of retrieved passages $m$ is set to 5.

\subsection{Implementation Details}
During our experiments, we utilize the base version of ALBERT(albert-base-v1) provided by huggingface\footnote{
    https://huggingface.co/transformers
} and initialize the parameters of the BERT encoding layer using pre-trained models that have been officially released by Google\footnote{
    https://github.com/google-research/albert
}. To retrieve knowledge, we use Sentence-BERT to obtain the embedding of the query sentence and passages. The size of embeddings of sentence is set to 768. For training our model, we utilize the ADAMW algorithm\citep{loshchilov2017decoupled} with a maximum learning rate of $2\times 10^{-5}$. We also employ a linear warmup for the first 1000 gradient steps, followed by a linear decrease of the learning rate for the next 9000 gradient steps. All experiments are conducted using an NVIDIA GeForce RTX 3090.

\begin{figure}[t]
    \centering
    \includegraphics[width=1\linewidth]{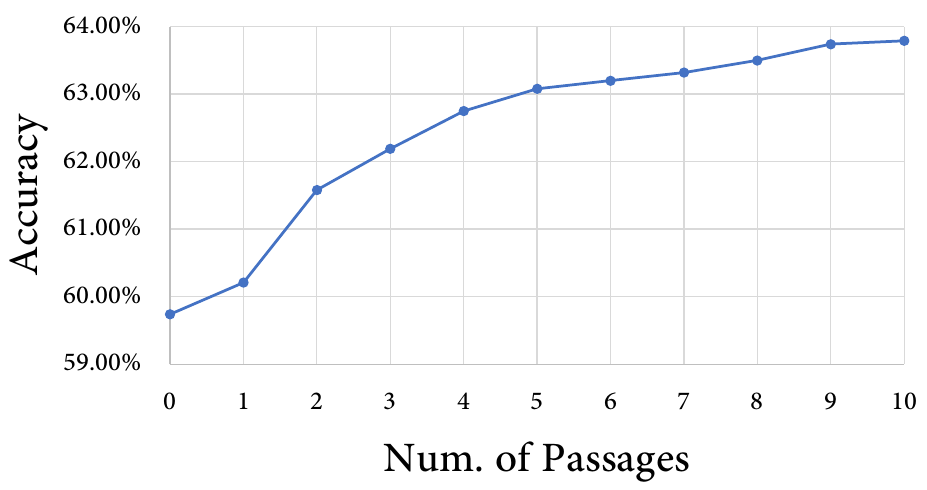}
    \caption{Performance of our model on the 600 meta-testing tasks of Amazon Review in the 1-shot scenario with the retrieval of different numbers of passages. It can be observed that our model is still effective even when the number of retrieved passages is 0. As the number of retrieved passages increases, there is a significant improvement in accuracy at the beginning. However, the performance improvement slows down after the number of retrieved passages reaches 5.}
    \label{fig:passages}
\end{figure}

\subsection{Experiment Results}
\paragraph{Baseline.}
We evaluate the effectiveness of our approach by comparing it with several baseline methods, including (1)\textbf{MAML}, an optimization-based method that trains models through learning to learn with gradients, (2)\textbf{ProtoNet}, which utilizes sample averages as class prototypes in a metric-based approach, (3)\textbf{InductNet}, a metric-based method that uses dynamic routing to learn class-wise representations, (4)\textbf{MatchingNet}, which uses metric-based attention to solve few-shot learning tasks, (5)\textbf{REGRAB} uses a global relation graph to assist few-shot classification, which captures the global relationship between relations, (6)\textbf{KGML} learns the representation from extracted sentence-specific knowledge graphs to enhance the representation of sentences.

\paragraph{Analysis.}
The main results on two datasets  are given in Table~\ref{tab:all}, it can be observed that our method significantly outperforms the baseline methods in all scenarios by utilizing both neural networks with parameters and non-parameterized knowledge obtained from a vast collection of texts. Moreover, in the 5-shot scenario, we observed greater enhancements for Huffpost as compared to Amazon Review, suggesting that the former is more responsive to external knowledge derived from Wikipedia. Compared to KGML, a method that also introduces external knowledge for meta-learning, our method significantly outperforms it in all scenarios, indicating that our multi-view passages fusion network better utilizes external knowledge.

Furthermore, the experimental results demonstrate the effectiveness of our method in few-shot learning.  Specifically, compared to the strongest baseline model, this method has increased the classification accuracy by 4.46\% and 0.97\% in the 1-shot and 5-shot scenarios, respectively, on the Amazon Review dataset. In the 1-shot and 5-shot scenarios on the Huffpost dataset, the accuracy has been improved by 2.33\% and 2.78\%, respectively. The improvement in performance suggests that the combination of parameterized neural networks and non-parameterized knowledge can effectively address the problem of data scarcity in few-shot learning. The success of our method suggests that utilizing both types of knowledge is a promising direction for future research in this field.

\subsection{Effectiveness of Introducing Knowledge}
To analyze the contribution and impact of external knowledge in our method, we conducted some ablation studies. Specifically, we varied the number of external passages retrieved on the query sentence and the performance is shown in Figure~\ref{fig:passages}. Notably, when the number of retrieved passages decreased to 0, our performance decreased significantly, but still outperformed the baseline method. This is attributed to the cross-attention mechanism that we introduced in meta-learning, which assigns different weights to each word in the query and achieves the characteristic of semantic preservation, thereby compensating for the difference in data between meta-training and meta-testing and alleviating the overfitting of neural networks to the training data.

As the retrieved information gradually increases, the performance improves significantly, which indicates that our multi-view passages fusion network effectively integrates the retrieved information for few-shot classification. When the number of retrieved messages exceeds 5, the speed of performance improvement begins to slow down, possibly due to the presence of duplicate and redundant information in the multiple retrieved messages. At this point, increasing the number of retrieved messages does not improve performance much, but rather affects computational efficiency. Therefore, we set the default number of retrieved messages to 5.

\subsection{Different pooling strategies}
As described in Section 4.2, we have different pooling strategies for the score matrix $\mathbf{S}$. We evaluated different pooling strategies (MEAN, MAX). Results are shown in Table~\ref{tab:pool}.

It can be observed that in all cases, the MEAN pooling strategy outperforms the MAX pooling strategy. This may be due to the fact that the MEAN pooling strategy can take into account all the passages and better integrate the knowledge conveyed by multiple passages.

\begin{table}[htbp]
  \centering
  \caption{Performance of the model under different pooling strategies.}
    \begin{tabular}{l|cc|cc}
    \toprule
    \multicolumn{1}{r}{} & \multicolumn{2}{c}{Amazon View} & \multicolumn{2}{c}{Huffpost} \\
    \midrule
          & 1-shot & 5-shot & 1-shot & 5-shot \\
    \midrule
    \multicolumn{1}{l}{Strategy} &       & \multicolumn{1}{c}{} &       &  \\
    \midrule
    MEAN  & \textbf{63.08}\% & \textbf{75.52}\% & \textbf{46.62}\% & \textbf{61.53}\% \\
    MAX   & 62.95\% & 75.38\% & 46.26\% & 61.50\% \\
    \bottomrule
    \end{tabular}%
  \label{tab:pool}%
\end{table}%

\begin{table}[htbp]
  \centering
  \caption{Results of meta-training time per task.}
    \begin{tabular}{c|c}
    \toprule
    \multicolumn{1}{c}{} & Training Time \\
    \midrule
    w/o passages & 0.213s \\
    with passages & 0.297s \\
    concat all & 1.531s \\
    \bottomrule
    \end{tabular}%
  \label{tab:time}%
\end{table}%

\subsection{Discussion of Computational Complexity}
An analysis of computational complexity was additionally conducted, and the meta-training time for each task was reported in Table~\ref{tab:time}. The experiments were carried out under 5-shot training of Amazon Review dataset. We also reported the training time of concatenating all passages into a single sentence instead of training in parallel through the passages fusion network, denoted as "\textit{concat all}".

The results indicate that we have made a good trade-off between computation time and how to utilize the introduced external retrieved information. First, for the default parallel processing method of retrieved passages in our model, it only slightly increases the training time, but brings significant benefits, namely, a significant improvement in accuracy, as shown in Figure~\ref{fig:passages}. However, if we concatenate all $m$ retrieved passages into one sentence, this will increase the training time several times, which is unacceptable to us.

\section{Conclusion}
In this paper, we propose RAML to address the issue of poor generalization performance in meta-learning, which introduces non-parametric external knowledge to solve this problem. Specifically, the external knowledge we introduce is different from previous parameterized networks for inference, which are prone to memorizing shallow features of the training set during training, making it difficult for the model's performance to generalize to the test set, especially when there is a large domain difference between the two sets. Our model can selectively use external knowledge and explicitly choose features that are beneficial for its generalization. Moreover, our passage fusion network uses cross-attention mechanism to fuse retrieved information in parallel, which endows it with the ability of semantic preservation, takes advantage of GPU parallel computing, and makes the retrieval process very efficient. Through extensive experiments, our method has been shown to outperform current SOTA meta-learning methods by a large margin. In future work, we will try to extend our method to the fields of computer vision and multimodal scenarios to see if it remains effective in those domains.

\bibliography{custem}
\end{document}